%% file: main.tex
\pdfoutput=1
\documentclass[11pt]{article}

\usepackage{acl}
\usepackage{times}
\usepackage{latexsym}
\usepackage[T1]{fontenc}
\usepackage[utf8]{inputenc}
\usepackage{microtype}
\usepackage{hyperref}
\usepackage{url}
\usepackage{booktabs}
\usepackage{amsfonts}
\usepackage{xcolor}
\usepackage{amsmath}
\usepackage{soul}
\usepackage{multirow}
\usepackage{xspace}
\usepackage{mwe}
\usepackage{listings}
\usepackage{lipsum}
\usepackage{subfigure}

\makeatletter
\def\@fnsymbol#1{\ensuremath{\ifcase#1\or \dagger\or *\or \ddagger\or
		\mathsection\or \mathparagraph\or \|\or **\or \dagger\dagger
		\or \ddagger\ddagger \else\@ctrerr\fi}}
\makeatother

\newcommand{\eg}{\emph{e.g.,}\xspace}

\newcommand{\ignore}[1]{}
\newcommand{\tabincell}[2]{\begin{tabular}{@{}#1@{}}#2\end{tabular}}

\DeclareRobustCommand{\textbox}{%
	\begingroup\normalfont
	\raisebox{-0.2em}{%
		\includegraphics[height=1.2em]{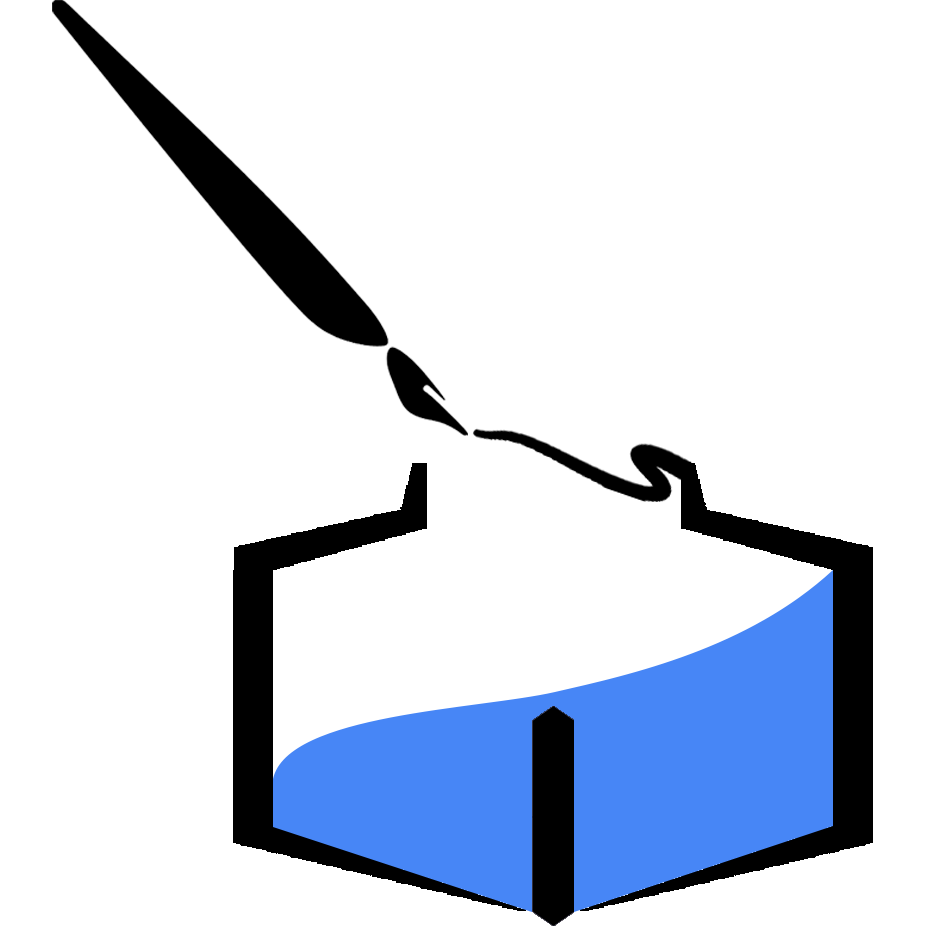}%
	}%
	\kern 0.4em%
	\endgroup
}

\title{\textbox TextBox 2.0: A Text Generation Library with \\ Pre-trained Language Models}

\author{
	Tianyi Tang\textsuperscript{\rm{1,5}}\thanks{\ \ Equal contribution.},~~
	Junyi Li\textsuperscript{\rm{1,3}}\footnotemark[1],~~
	Zhipeng Chen\textsuperscript{\rm{2}}\footnotemark[1],~~
	Yiwen Hu\textsuperscript{\rm{2}},~~
	Zhuohao Yu\textsuperscript{\rm{2}}, \\
	\textbf{Wenxun Dai}\textsuperscript{\rm{4}}\textbf{,~~}
	\textbf{Zican Dong}\textsuperscript{\rm{1}}\textbf{,~~}
	\textbf{Xiaoxue Cheng}\textsuperscript{\rm{1}}\textbf{,~~}
	\textbf{Yuhao Wang}\textsuperscript{\rm{1}}\textbf{,~~} \\
	\textbf{Wayne Xin Zhao}\textsuperscript{\rm{1,5}\thanks{\ \ Corresponding author}\ }\textbf{,~~}
	\textbf{Jian-Yun Nie}\textsuperscript{\rm{3}}, \and
	\textbf{Ji-Rong Wen}\textsuperscript{\rm{1,2,5}} \\
	\textsuperscript{1}Gaoling School of Artificial Intelligence, Renmin University of China \\
	\textsuperscript{2}School of Information, Renmin University of China \\
	\textsuperscript{3}DIRO, Université de Montréal \quad \textsuperscript{4}Xidian University \\
	\textsuperscript{5}Engineering Research Center of Next-Generation Intelligent Search and Recommendation, MOE\\
	\texttt{steventianyitang@outlook.com lijunyi@ruc.edu.cn batmanfly@gmail.com} \\ 
}

\begin{document}
\maketitle
\begin{abstract}
To facilitate research on text generation, this paper presents a comprehensive and unified library, \textbf{TextBox 2.0}, focusing on the use of pre-trained language models (PLMs). 
To be \emph{comprehensive}, our library covers $13$ common text generation tasks and their corresponding $83$ datasets and further incorporates $45$ PLMs covering general, translation, Chinese, dialogue, controllable, distilled, prompting, and lightweight PLMs. We also implement $4$ efficient training strategies and provide $4$ generation objectives for pre-training new PLMs from scratch.
To be \emph{unified}, we design the interfaces to support the entire research pipeline (from data loading to training and evaluation), ensuring that each step can be fulfilled in a unified way. Despite the rich functionality, it is easy to use our library, either through the friendly Python API or command line. 
To validate the effectiveness of our library, we conduct extensive experiments and exemplify four types of research scenarios.
The project is released at the link: \url{https://github.com/RUCAIBox/TextBox#2.0}.
\end{abstract}

\input{sec-intro} 
\input{sec-design} 
\input{sec-usage} 
\input{sec-exp} 

\section{Conclusion}
This paper presented \textbf{TextBox 2.0}, a comprehensive and unified library for conducting research on PLM-based text generation. Our library makes significant extensions in three major aspects, namely generation tasks ($13$ tasks and $83$ datasets), generation models ($45$ PLMs), and training strategies (\eg distributed data parallel and efficient decoding). Results from extensive test experiments demonstrate that our library can accurately reproduce existing models. Besides, we also provide a series of utility tools to better analyze and explore the generated results. 
To summarize, our library can be very useful to facilitate text generation research, and our team will improve this library with regular updates.

\section*{Acknowledgement}
This work was partially supported by Beijing Natural Science Foundation under Grant No. 4222027, and Beijing Outstanding Young Scientist Program under Grant No. BJJWZYJH012019100020098. Xin Zhao is the corresponding author.

\bibliography{ref}
\bibliographystyle{acl_natbib}

\end{document}

%% file: sec-intro.tex
\section{Introduction}

Text generation, aiming to generate human-like texts on demand, has been a fundamental technique in many text applications, such as machine translation~\cite{mt_survey}, text summarization~\cite{sum_survey}, and dialogue system~\cite{dia_survey}. Recently, pre-trained language models (PLMs) such as BART~\citep{bart} have been the mainstream approach to developing effective text generation models. With the great advances in text generation, it has become increasingly important to reproduce, develop, and compare various text generation models in a reliable, flexible, and unified way.





Considering the rapid progress of PLMs on text generation, in this paper, we present a significant extension of a previously released text generation library, \emph{TextBox 1.0}~\cite{textbox}, called \textbf{TextBox 2.0}. Different from TextBox 1.0 and other text generation libraries~\cite{parlai,opennmt,texygen} (mostly including classical models based on recurrent neural networks or generative adversarial networks), this extension mainly focuses on building a comprehensive and unified framework for better supporting PLM-based text generation models.   
Although some libraries (\eg Fairseq~\cite{fairseq} and Hugging Face~\cite{huggingface}) also include PLMs, they are designed for performing myriad NLP tasks (only considering a few text generation tasks). Moreover, they don't maintain a complete evaluation pipeline (\eg data loading, training, inference, and evaluation) specially designed for text generation. Thus, it is not fully suited for developing and evaluating text generation models in a unified way.

\begin{table*}
	\small
	\renewcommand*{\arraystretch}{1.12}
	\begin{tabular}{p{0.085\textwidth}p{0.31\textwidth}p{0.525\textwidth}}
		\toprule
	\textbf{Aspects}	& \tabincell{c}{\textbf{TextBox 1.0}} & \tabincell{c}{\textbf{TextBox 2.0}} \\
		\midrule
		\tabincell{l}{\textbf{Tasks} \\ \textcolor{blue}{6 \emph{v.s.} 13}} & 
			\tabincell{l}{Summarization, translation, dialogue, \\unconditional generation, attribute-\\to-text generation, poem generation} & 
			\tabincell{l}{Summarization, translation, dialogue, data-to-text, question genera-\\tion, question answering, story generation, commonsense generation,\\ Chinese generation, paraphrase, style transfer and simplification} \\
		\midrule[0.3pt]
		\tabincell{l}{\textbf{Models} \\ \textcolor{blue}{6 \emph{v.s.} 45}} & 
			\tabincell{l}{\textbf{VAE}: LSTMVAE, CNNVAE, CVAE, \\ HybridVAE \\ 
						  \textbf{GAN}: SeqGAN, TextGAN, RankGAN, \\ MaliGAN, LeakGAN, MaskGAN \\ 
						  \textbf{PLM}: GPT-2, XLNet, BERT2BERT, T5, \\ BART, ProphetNet \\
						  \textbf{Seq2Seq}: RNN, Transformer, Attr2Seq, \\ Context2Seq, HRED}  & 
			\tabincell{l}{\textbf{General}: GPT-2, BERT2BERT, BART, T5, ProphetNet, GPT, GPT-\\ Neo, OPT, UniLM, MASS, PEGASUS, MVP, Bigbird, LED \\
						  \textbf{Translation}: mBART, mT5, Marian, M2M 100, NLLB, XLM \\
						  \textbf{Chinese}: CPM, CPT, Chinese-BART, Chinese-T5, Chinese-GPT2 \\
						  \textbf{Dialogue}: Blenderbot and DialoGPT \\
						  \textbf{Controllable}: CTRL and PPLM \\
						  \textbf{Distilled}: DistilGPT2 and DistilBART \\
						  \textbf{Prompting}: PTG and Context-Tuning \\
                          \textbf{Lightweight}: Adapter, Prefix-tuning, Prompt tuning, LoRA, BitFit, \\ P-Tuning v2
		} \\
		\midrule[0.3pt]
		\textbf{Training Strategies} & \multirow{2}{*}{Distributed data parallel} & Distributed data parallel, efficient decoding, hyper-parameter optimization, repeated experiments, pre-training objectives \\
		\bottomrule
	\end{tabular}
	\caption{Comparison of TextBox 1.0 and TextBox 2.0. We also present a comparison of the numbers of \emph{tasks} and pre-trained \emph{models} between the two versions.}
	\label{tb-comparison}
\end{table*}

In order to better facilitate research on text generation, \textbf{TextBox 2.0} introduces a series of new  features for supporting the use of PLMs, which can be summarized into three major aspects:  

\textbullet~\textit{Generation Tasks}: Our library supports $13$ commonly studied text generation tasks (\eg translation and story generation) and their corresponding $83$ datasets, including most of the existing mainstream tasks and datasets for research. We reorganize these datasets so that they are framed in a unified text-to-text format. Users can simply set the dataset via the command line or configuration file without additional preprocessing efforts.  

\textbullet~\textit{Generation Models}: As a key contribution, our library incorporates $45$ pre-trained language models, covering the categories of general, translation, Chinese, dialogue, controllable, distilled, prompting, and lightweight models (modules). We unify the interface to use existing PLMs and incorporate new PLMs, and it is convenient to run different PLMs for a specified task in our library. 
We also provide a standard way to compare these models and analyze the generated results. 

\textbullet~\textit{Training Strategies}: To support the optimization of PLMs, we provide four efficient and robust training strategies (\eg efficient decoding) and four pre-training objectives (\eg denoising auto-encoding) for text generation. These strategies make optimizing text generation models more efficient and reliable. Users can either pre-train a new model from scratch or fine-tune a pre-trained model for research purposes.

As another merit, TextBox 2.0 has been largely aligned with our previous survey on PLM-based text generation~\cite{tg_survey} in terms of task, model, and training. It will be meaningful for both research beginners and experts to learn and explore text generation models with the survey and accordingly supporting libraries. 

To summarize, TextBox 2.0 has contributed a significant addition to the previous version (see Table~\ref{tb-comparison} for a detailed comparison) to better support the use of PLMs for text generation. It implements and maintains a unified way to conduct research on text generation with $45$ included models, covering $13$ tasks, and $83$ datasets. We also perform extensive test experiments, and these results show that TextBox 2.0 can produce very competitive performance compared to the original implementations. 


%% file: sec-design.tex
\section{Library Design}


In order to facilitate PLM-based text generation research, TextBox 2.0 has introduced various new features, mainly from three aspects: \textit{generation tasks}, \textit{generation models}, and \textit{training strategies}.


\subsection{Generation Tasks} \label{sec:task}
Since there are various text generation applications, we include $13$ widely studied tasks and collect the corresponding $83$ datasets.  

\paragraph{Tasks.} These $13$ tasks in TextBox 2.0 include text summarization, machine translation, open-ended dialogue system, data-to-text generation, question generation, question answering, story generation, task-oriented dialogue system, commonsense generation, paraphrase generation, text style transfer, and text simplification. Besides these English-centric tasks, we also include Chinese generation tasks. Existing PLM-based libraries such as Hugging Face~\cite{huggingface} are focused on performing extensive NLP tasks and only consider a few text generation tasks (mainly text summarization and machine translation), which are not comprehensive for text generation research. 


\paragraph{Datasets.} For each task, we collect widely-used datasets and reorganize them in a unified text-to-text format. In total, we include $83$ datasets, and report their details on the page\footnote{\url{https://github.com/RUCAIBox/TextBox\#dataset}}, including the dataset description, basic statistics, and training/validation/testing samples. In addition, we build a leaderboard for each dataset by collecting the automatic results and generated texts of the latest research. It is convenient for users to quickly learn about the baselines and their results. We also encourage community users to collaboratively maintain the leaderboard and submit their model results.

\paragraph{Metrics.} To conduct evaluations with these tasks and datasets, TextBox 2.0 supports four categories of automatic metrics: 
(1) lexical metrics, such as BLEU~\citep{bleu} and ROUGE~\citep{rouge}, to measure the $n$-gram overlap between generated texts and golden texts;
(2) semantic metrics, such as BERTScore~\citep{bertscore} and style strength~\citep{sc_bleu}, to compare the texts at sentence level;
(3) diversity metrics, such as Distinct~\citep{distinct} and Self-BLEU~\cite{texygen}, to evaluate the lexical diversity of generated texts;
(4) accuracy metrics, such as exact match~\citep{squad} and inform~\citep{multiwoz}, to calculate the precision of important phrases.
In total, we include $12$ general metrics and $5$ task-specific metrics\footnote{\url{https://github.com/RUCAIBox/TextBox\#evaluation}}. 

Besides the analysis using automatic metrics, TextBox 2.0 provides several visualization tools to explore and analyze the generated texts in various dimensions~\citep{explainaboard,sentspace}. For instance, Figure~\ref{fig:analysis} shows how it offers new insights to improve summarization tasks (details can be found in Section~\ref{sec:visual}). 


\subsection{Generation Models}
To support the rapid progress of PLMs on text generation, TextBox 2.0 incorporates $45$ PLMs\footnote{\url{https://github.com/RUCAIBox/TextBox\#model}} and aims to build a unified and standardized framework based on PLMs. 
We list some included models as  follows:



\textbullet~\textbf{General PLMs}: GPT-2~\citep{gpt2} and BART~\citep{bart};

\textbullet~\textbf{Translation PLMs}: mBART~\citep{mbart} and XLM~\citep{xlm};

\textbullet~\textbf{Chinese PLMs}: CPM~\citep{cpm} and CPT~\citep{cpt};

\textbullet~\textbf{Dialogue PLMs}: DialoGPT~\citep{dialogpt} and Blenderbot~\citep{blenderbot}; 

\textbullet~\textbf{Controllable PLMs}: CTRL~\citep{ctrl} and PPLM~\citep{pplm};

\textbullet~\textbf{Distilled PLMs}: DistilGPT2~\citep{distilbert} and DistilBART~\citep{distilbart};

\textbullet~\textbf{Prompting PLMs}: PTG~\citep{ptg} and Context-Tuning~\citep{context-tuning};

\textbullet~\textbf{Lightweight modules}: Adapter~\cite{adapter} and Prefix-tuning~\cite{prefix-tuning}.

The wide coverage of PLMs makes it possible to deal with different text generation tasks using TextBox 2.0. 
For example, to perform specific tasks such as dialogue system, users can adopt task-specific PLMs such as DialoGPT; to deal with Chinese generation tasks, users can adopt CPT. In resource-constrained situations, lightweight PLMs such as prefix-tuning can be a good choice.


\subsection{Training Strategies} \label{sec-training}
TextBox 2.0 provides four pre-training objectives to help users pre-train a model from scratch, including language modeling~\cite{gpt2}, masked sequence-to-sequence modeling~\cite{mass}, denoising auto-encoding~\cite{bart}, and masked span prediction~\cite{t5}. These pre-training tasks can also be utilized for domain-adaptive pre-training and task-adaptive pre-training~\citep{dapt&tapt} to tailor existing PLM to the domain of a target task.

Also, TextBox 2.0 provides four useful training methods for improving the optimization of PLMs. It supports distributed data parallel to implement models on multiple GPUs and machines to improve the efficiency of training. We incorporate \emph{Accelerate}~\citep{accelerate} to support distributed training with a simple API. To further accelerate the decoding efficiency, we integrate FastSeq~\citep{fastseq} to optimize the decoding process by attention cache optimization, repeated $n$-gram detection, and asynchronous parallel I/O.

Moreover, TextBox 2.0 enables users to adjust and select hyper-parameters automatically. Based on the library Hyperopt~\citep{hyperopt}, users just need to set the parameter range and search methods, and then the optimal hyper-parameters and corresponding results will return. It is useful for PLMs to search for hyper-parameters such as batch size and learning rate. Our library also supports performing repeat experiments using different random seeds in one command line, which is especially useful to alleviate randomness especially under few-shot settings. 


%% file: sec-usage.tex
\begin{figure*}[!t]
	\centering
	\includegraphics[width=1.0\textwidth]{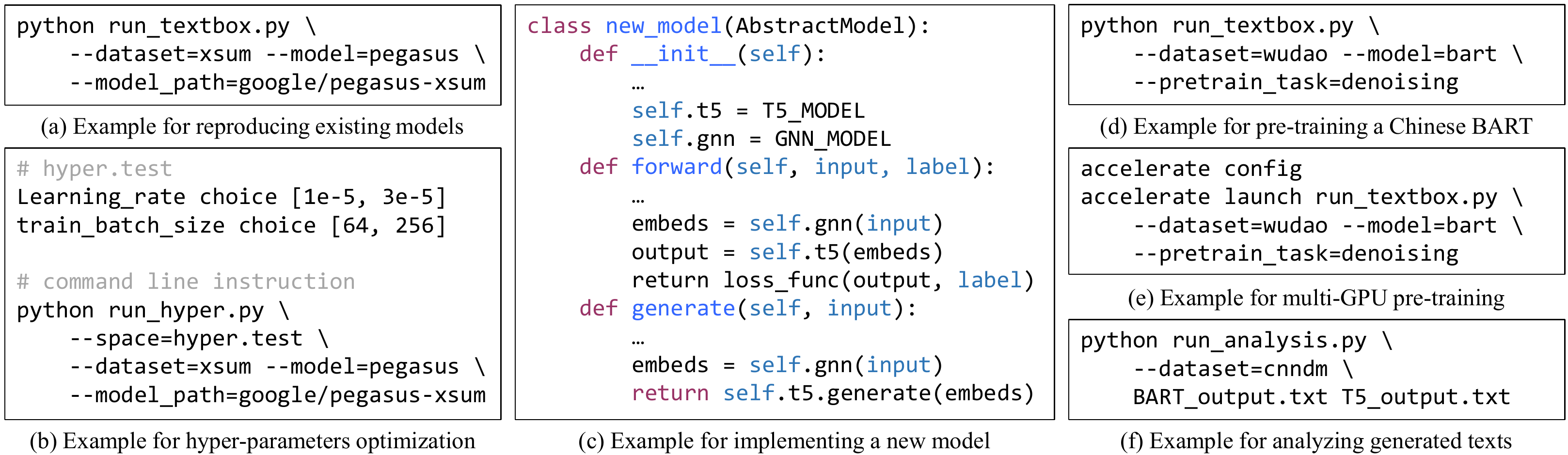}
	\caption{Example usage of our TextBox 2.0.}
	\label{fig-usage}
\end{figure*}

\section{Library Usage}

In this section, we introduce how to use our library in four different kinds of research scenarios by showing the example codes.


\paragraph{Reproducing existing models.}
TextBox 2.0 includes various PLMs and supports many text generation tasks and datasets. It is convenient for users to quickly run existing PLMs and reproduce results for each dataset. In particular, users only need to specify the dataset and model by setting the configurations \texttt{dataset}, \texttt{model}, and \texttt{model\_path}, within a simple command line. 

Figure~\ref{fig-usage}(a) presents an example to fine-tune PEGASUS~\citep{pegasus} on XSum~\cite{xsum} dataset. Moreover, TextBox 2.0 enables users to conduct hyper-parameter optimization by only providing a list of possible values. Figure \ref{fig-usage}(b) shows an example that automatically adjusts the hyper-parameters \texttt{learning\_rate} and \texttt{batch\_size} from the ranges $[1\times 10^{-5}, 3\times 10^{-5}]$ and $[64, 256]$, respectively.


\paragraph{Implementing a new model.}
Since TextBox 2.0 builds a unified pipeline for text generation research, users only need to define a new model class without considering other procedures to implement a new model. Specially, users should first inherit from our base model class \texttt{AbstractModel} before specifying three specific model functions: (1) \texttt{\_\_init\_\_()}: this function initializes the architectures and parameters of the model; (2) \texttt{forward()}: this function is used to calculate the loss for optimization during training; (3) \texttt{generate()}: this function generates texts based on input during inference.

Figure \ref{fig-usage}(c) presents an example of implementing a new model for the KG-to-text generation task . 
In this example, the model adopts a graph neural network (GNN) to encode KG and then uses T5~\citep{t5} to generate texts. We first define the GNN and T5 models in the \texttt{\_\_init\_\_()} function. Then, we use GNN to encode KG to embeddings as the input of T5 and compute the loss according to target labels in the \texttt{forward()} function. Finally, we use a similar process to generate text in the \texttt{generate()} function.

\paragraph{Pre-training a new model.}
In TextBox 2.0, we provide several pre-training objectives for users to pre-train new models from scratch. Specifically, users just need to specify the pre-training task, pre-training corpus, and architecture by setting \texttt{pretrain\_task}, \texttt{dataset}, and \texttt{model}. Figure \ref{fig-usage}(d) shows an example that pre-trains a Chinese BART on the WuDaoCorpora~\citep{wudao} using the denoising pre-training objective. 

To improve the pre-training efficiency, TextBox 2.0 supports distributed data parallel and efficient decoding (Section~\ref{sec-training}). Figure \ref{fig-usage}(e) shows an illustrative example of how users can use the \texttt{accelerate} command to set  configurations of multiple devices and launch the training code.

\paragraph{Analyzing generated results.} 
Besides simply obtaining the evaluation results, our library provides several visualization analysis mechanisms to perform deep analysis on the generated results of models. For example, we support the use of the statistical chart to analyze the mean and standard deviation scores for different sentence lengths. These methods can help users learn about the advantages and disadvantages of different models in a detailed comparison. Figure \ref{fig-usage}(f) shows an example of how to run the analysis using a simple command line and the results can be found in Figure~\ref{fig:analysis}. This example compares the generated texts of BART and T5 on the CNN/DailyMail dataset.

%% file: sec-exp.tex
\begin{table*}[htbp!]
	\resizebox{1\textwidth}{!}{
		\begin{tabular}{rlllllllllll}
			\toprule
			& \multicolumn{3}{c}{\textbf{Text Summarization}} & \multicolumn{3}{c}{\textbf{Text Simplification}} & \multicolumn{3}{c}{\textbf{Chinese Generation}} & \multicolumn{2}{c}{\textbf{Translation}} \\ 
			\cmidrule(lr){2-4} \cmidrule(lr){5-7} \cmidrule(lr){8-10} \cmidrule(lr){11-12} 
			& \multicolumn{1}{c}{R-$1$} & \multicolumn{1}{c}{R-$2$} & \multicolumn{1}{c}{R-L} & \multicolumn{1}{c}{B-$4$} & \multicolumn{1}{c}{ME} & \multicolumn{1}{c}{R-$2$} & \multicolumn{1}{c}{LCSTS} & \multicolumn{1}{c}{CSL} & \multicolumn{1}{c}{ADGEN} & \multicolumn{1}{c}{En$\rightarrow$Ro} & \multicolumn{1}{c}{Ro$\rightarrow$En} \\ 
			\midrule
			BART  & 44.16\textsuperscript{\textit{a}} & 21.28 & 40.90 & 88.30\textsuperscript{\textit{b}} & 55.60 & 86.10 & 40.60\textsuperscript{\textit{c}} & 64.20 & 10.00 & 37.70\textsuperscript{\textit{d}} & 37.80 \\
			BART (ours)  & 44.47\textsubscript{\textsubscript{0.10}} & 21.50\textsubscript{\textsubscript{0.14}} & 41.35\textsubscript{\textsubscript{0.08}} & 90.81\textsubscript{\textsubscript{0.24}} & 57.58\textsubscript{\textsubscript{0.19}} & 83.36\textsubscript{\textsubscript{0.07}} & 42.96\textsubscript{\textsubscript{0.18}} &  64.34\textsubscript{\textsubscript{0.63}} &
			10.20\textsubscript{\textsubscript{0.15}} &
			37.20\textsubscript{\textsubscript{0.17}} & 37.48\textsubscript{\textsubscript{0.31}} \\
			
			\midrule
			& \multicolumn{3}{c}{\textbf{Data-to-text Generation}} & \multicolumn{3}{c}{\textbf{Commonsense Generation}} & \multicolumn{3}{c}{\textbf{Question Generation}} & \multicolumn{2}{c}{\textbf{QA}} \\ 
			\cmidrule(lr){2-4} \cmidrule(lr){5-7} \cmidrule(lr){8-10} \cmidrule(lr){11-12} 
			& \multicolumn{1}{c}{B-$4$} & \multicolumn{1}{c}{ME} & \multicolumn{1}{c}{R-L} & \multicolumn{1}{c}{B-$4$} & \multicolumn{1}{c}{CIDEr} & \multicolumn{1}{c}{SPICE} & \multicolumn{1}{c}{B-$4$} & \multicolumn{1}{c}{ME} & \multicolumn{1}{c}{R-L} & \multicolumn{1}{c}{F1} & \multicolumn{1}{c}{EM} \\ 
			\midrule
			BART  & 64.55\textsuperscript{\textit{e}} & 46.51 & 75.13 & 27.50\textsuperscript{\textit{f}} & 14.12 & 30.00 & 22.00\textsuperscript{\textit{g}} & 26.40 & 50.30 & 91.56\textsuperscript{\textit{h}} & 84.23 \\
			BART (ours)  & 67.33\textsubscript{\textsubscript{0.06}} & 47.78\textsubscript{\textsubscript{0.07}} & 76.83\textsubscript{\textsubscript{0.04}} & 28.18\textsubscript{\textsubscript{0.45}} & 12.98\textsubscript{\textsubscript{0.13}} & 33.00\textsubscript{\textsubscript{0.40}} & 25.08\textsubscript{\textsubscript{0.13}} & 26.73\textsubscript{\textsubscript{0.18}} & 52.55\textsubscript{\textsubscript{0.07}} & 93.04\textsubscript{\textsubscript{0.08}} & 86.44\textsubscript{\textsubscript{0.21}} \\
			
			\midrule
			& \multicolumn{4}{c}{\textbf{Open-ended Dialogue System}} & \multicolumn{4}{c}{\textbf{Task-oriented Dialogue System}} & \multicolumn{3}{c}{\textbf{Story Generation}}           \\ 
			\cmidrule(lr){2-5} \cmidrule(lr){6-9} \cmidrule(lr){10-12}
			& \multicolumn{1}{c}{B-$1$} & \multicolumn{1}{c}{B-$2$} & \multicolumn{1}{c}{D-$1$} & \multicolumn{1}{c}{D-$2$} & \multicolumn{1}{c}{B-$4$} & \multicolumn{1}{c}{Success} & \multicolumn{1}{c}{Inform} & \multicolumn{1}{c}{Comb.} & \multicolumn{1}{c}{B-$1$} & \multicolumn{1}{c}{B-$2$} & \multicolumn{1}{c}{D-$4$} \\  
			\midrule
			BART  & 49.90\textsuperscript{\textit{g}} & 40.00 & 1.30 & 8.00 & 17.89\textsuperscript{\textit{i}} & 74.91 & 84.88 & 97.78 & 30.70\textsuperscript{\textit{j}} & 13.30 & 69.90\\
			BART (ours)  & 49.58\textsubscript{\textsubscript{1.12}} & 39.24\textsubscript{\textsubscript{0.90}} & 1.44\textsubscript{\textsubscript{0.09}} & 8.89\textsubscript{\textsubscript{0.57}} & 20.17\textsubscript{\textsubscript{0.63}} & 75.40\textsubscript{\textsubscript{1.22}} & 84.40\textsubscript{\textsubscript{1.15}} & 100.07\textsubscript{\textsubscript{0.53}} & 33.79\textsubscript{\textsubscript{0.13}} & 15.78\textsubscript{\textsubscript{0.21}} & 78.76\textsubscript{\textsubscript{2.15}} \\
			
			\midrule
			& \multicolumn{5}{c}{\textbf{Paraphrase Generation}} & \multicolumn{3}{c}{\textbf{Style Transfer (E\&M)}} & \multicolumn{3}{c}{\textbf{Style Transfer (F\&R)}} \\ 
			\cmidrule(lr){2-6} \cmidrule(lr){7-9} \cmidrule(lr){10-12}
			& \multicolumn{1}{c}{B-$4$} & \multicolumn{1}{c}{ME} & \multicolumn{1}{c}{R-$1$} & \multicolumn{1}{c}{R-$2$} & \multicolumn{1}{c}{R-L} & \multicolumn{1}{c}{B-$4$} & \multicolumn{1}{c}{Acc.} & \multicolumn{1}{c}{HM} & \multicolumn{1}{c}{B-$4$} & \multicolumn{1}{c}{Acc.} & \multicolumn{1}{c}{HM} \\
			\midrule
			BART & 47.30\textsuperscript{\textit{k}} & 49.70 & 73.30 & 54.10 & 75.10 & 76.50\textsuperscript{\textit{l}} & 92.90 & 83.90 & 79.30 & 92.00 & 85.20 \\
			BART (Ours) & 48.35\textsubscript{\textsubscript{0.70}} & 50.60\textsubscript{\textsubscript{0.49}} & 74.16\textsubscript{\textsubscript{0.47}} & 55.25\textsubscript{\textsubscript{0.74}} & 75.84\textsubscript{\textsubscript{0.42}} & 76.93\textsubscript{\textsubscript{0.55}} & 94.37\textsubscript{\textsubscript{0.87}} & 84.74\textsubscript{\textsubscript{0.05}} & 80.11\textsubscript{\textsubscript{0.29}} & 92.29\textsubscript{\textsubscript{0.37}} & 85.77\textsubscript{\textsubscript{0.10}} \\
			\bottomrule
	\end{tabular}}
	\caption{The results of BART on thirteen tasks from the original papers and our TextBox 2.0. QA is short for question answering. B, R, D, ME, EM, HM, Acc., and Comb. denote BLEU, ROUGE, Distinct, METEOR, exact match, harmonic mean, accuracy, and combined score, respectively. LCSTS, CSL, ADGEN, and En$\leftrightarrow$Ro are evaluated using the R-L, R-L, B-$4$, and B-$4$ metrics, respectively. 
		\quad \textsuperscript{\textit{a}}\citep{bart} \quad \textsuperscript{\textit{b}}\citep{gem} \quad \textsuperscript{\textit{c}}\citep{cpt} \quad \textsuperscript{\textit{d}}\citep{mbart} \quad \textsuperscript{\textit{e}}\citep{jointgt} \quad \textsuperscript{\textit{f}}\citep{cg} \quad \textsuperscript{\textit{g}}\citep{glge} \quad \textsuperscript{\textit{h}}\citep{qa} \quad \textsuperscript{\textit{i}}\citep{mintl} \quad \textsuperscript{\textit{j}}\citep{hint} \quad \textsuperscript{\textit{k}}\citep{aesop}\quad \textsuperscript{\textit{l}}\citep{sc_bleu}}
	\label{tab:results}
\end{table*}

\section{Experiments}

In this section, we conduct extensive experiments to verify the generation abilities of TextBox 2.0.

\subsection{Result Reproduction} \label{sec:res}
As an open-source library, TextBox 2.0 should be able to reproduce the results of existing work effectively. To verify this, we select a number of widely-used datasets for each task (introduced in Section~\ref{sec:task}) and compare the results conducted by TextBox 2.0 with those in the original papers. We totally evaluate $13$ tasks using $14$ datasets, including CNN/DailyMail~\citep{cnndm}, WikiAuto + Turk~\citep{glge}, LCSTS~\citep{lcsts}, CSL\footnote{\url{https://github.com/CLUEbenchmark/CLGE}}, ADGEN~\citep{adgen}, WMT 16
English-Romanian (En$\leftrightarrow$Ro)~\citep{wmt}, WebNLG 2.1~\citep{webnlg}, CommonGen~\citep{cg}, SQuAD~\citep{squad}, PersonaChat~\citep{pc}, MultiWOZ 2.0~\citep{mw}, ROCStories~\citep{roc}, GYAFC (E\&M and F\&R)~\citep{gyafc}, and Quora~\citep{quora}.

Since BART is the prevalent PLM for text generation, we endeavor to reproduce existing works with BART\textsubscript{\textsc{large}}\footnote{For translation tasks, we utilize mBART-CC25~\cite{mbart}. For Chinese generation tasks, we utilize Chinese BART\textsubscript{\textsc{large}}~\cite{cpt}.}. For all experiments, we employ the sequence-to-sequence cross-entropy loss with a label smoothing factor of $0.1$ as the objective function. We optimize the model using AdamW~\cite{adamw} with a constant learning rate of $3 \times 10^{-5}$. The accumulated batch size is set to $192$. During inference, we apply beam search with a beam size of $5$ and no-repeat $n$-gram size of $3$. To reduce randomness, we report the mean and standard deviation of our results based on three random seeds: $2020$, $2021$, and $2022$. All codes are implemented in PyTorch 1.11.0 with NVIDIA GeForce RTX 3090 24GB. 

To conduct these experiments, we only need to run the script shown in Figure~\ref{fig-usage} (a) with different \texttt{dataset} names. 
As shown in Table~\ref{tab:results}, our TextBox 2.0 can faithfully reproduce the results reported in existing work. Remarkably, our library achieves better performances than original works on $37$ of the $44$ metrics evaluated. It might be because we adopt optimization strategies such as label smoothing and large batch sizes. 


\begin{figure*}[t]
	\centering
	\subfigure[Leaderboard of CNN/DailyMail]{
		\centering
		\includegraphics[width=0.31\textwidth]{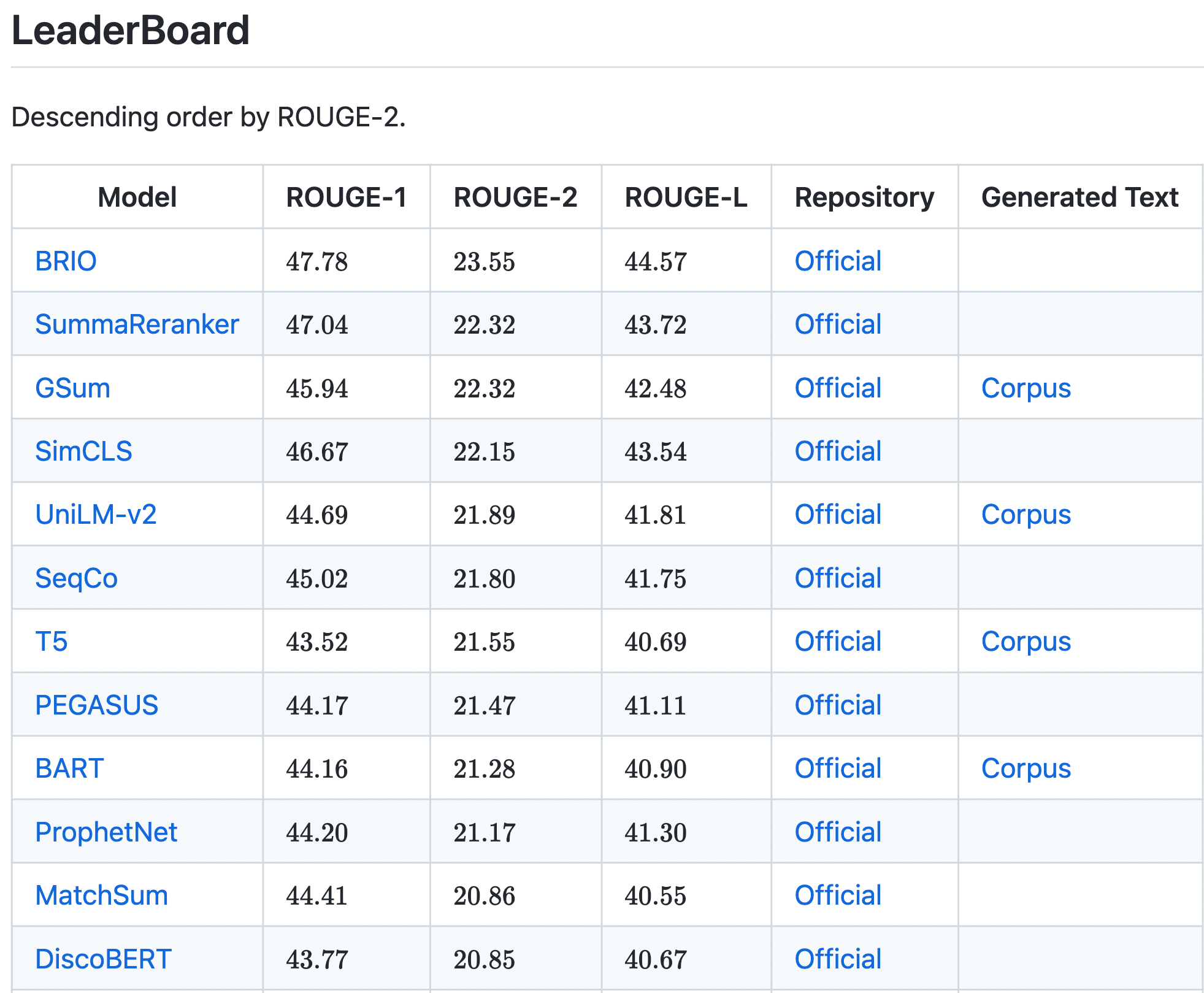}
	}
	\subfigure[ROUGE-L scores of BART and T5 for different input lengths]{
		\centering
		\includegraphics[width=0.31\textwidth]{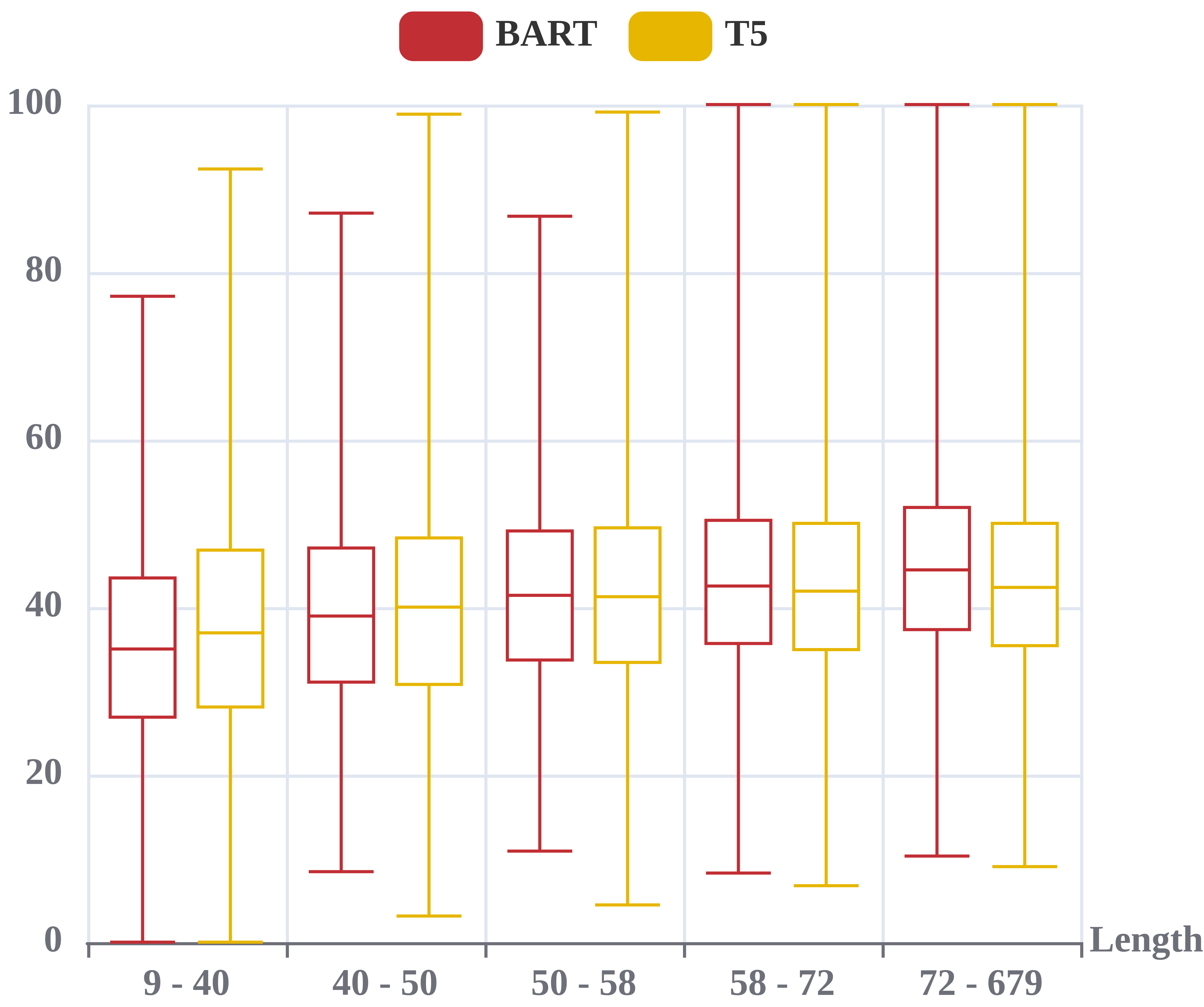}
	}
	\subfigure[N-gram overlap of target and generated texts with input document]{
		\centering
		\includegraphics[width=0.31\textwidth]{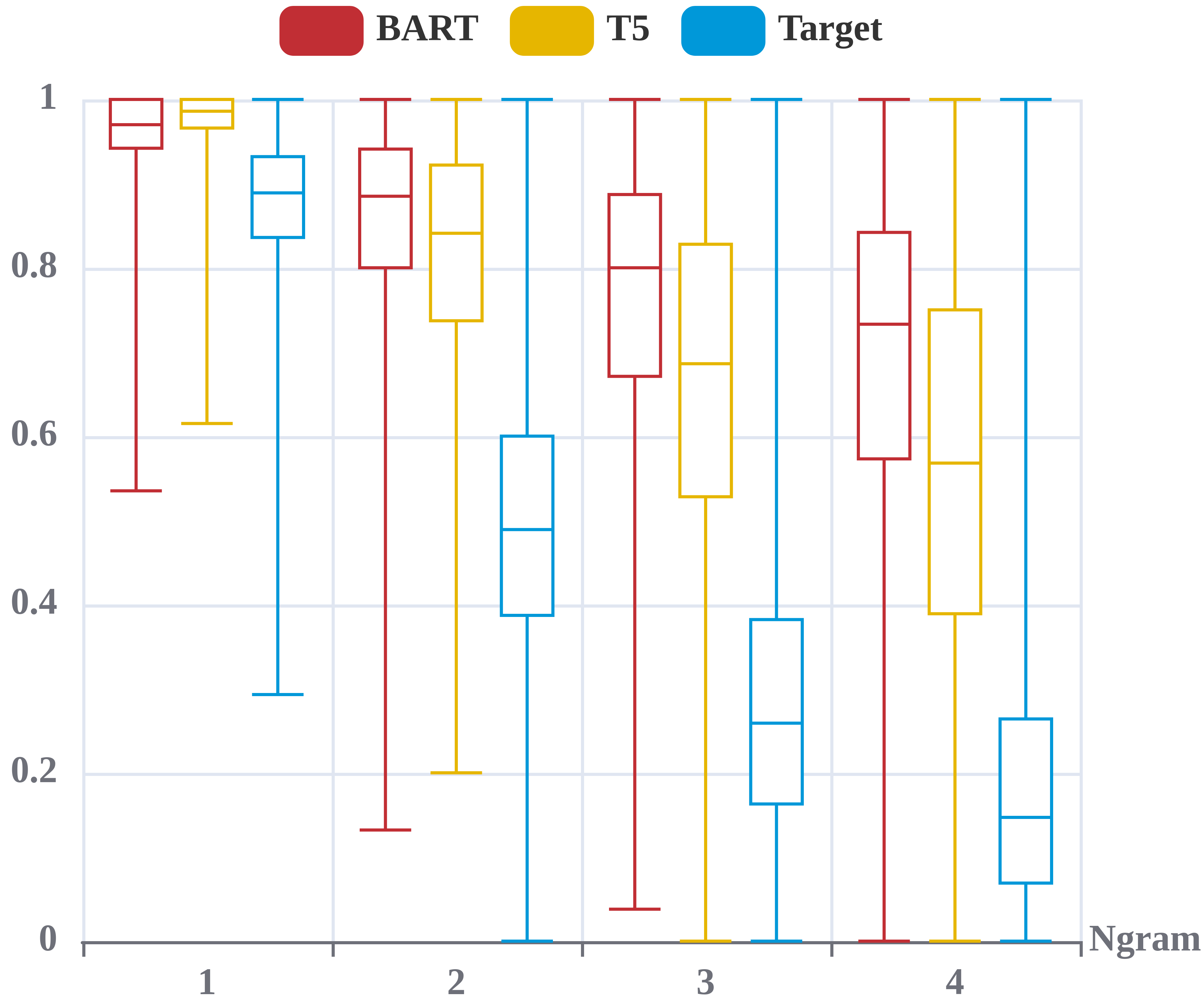}
	}
	\centering
	\caption{The partial visualization analysis on CNN/DailyMail dataset. The whole one can be found at \url{https://github.com/RUCAIBox/TextBox/blob/2.0.0/asset/example-analysis.html}.}
	\label{fig:analysis}
\end{figure*}

\begin{table}[t]
	\resizebox{1\columnwidth}{!}{
		\begin{tabular}{rccc}
			\toprule
			\multirow{2}{*}{\textbf{Library}} & \textbf{Preparation} & \textbf{Training} & \textbf{Generation} \\
			& (minutes)            & (minutes)         & (minutes)           \\
			\midrule
			\textbf{Fairseq}                  & 2.93\textsubscript{\textsubscript{0.02}} & 410.05\textsubscript{\textsubscript{8.86}} & 79.24\textsubscript{\textsubscript{1.50}} \\
			\textbf{Hugging Face}             & 4.02\textsubscript{\textsubscript{0.12}} & 416.25\textsubscript{\textsubscript{4.47}} & 75.69\textsubscript{\textsubscript{2.53}} \\
			\textbf{TextBox 2.0}                  & 3.81\textsubscript{\textsubscript{0.14}} & 393.99\textsubscript{\textsubscript{5.09}} & 27.05\textsubscript{\textsubscript{1.03}} \\
			\bottomrule
	\end{tabular}}
	\caption{Efficiency comparison of three libraries for BART\textsubscript{\textsc{large}} fine-tuned on CNN/DailyMail. The preparation stage consists of configuration loading, text tokenization, and necessary initialization options. The training stage takes time for fine-tuning on the training set in one epoch. The generation stage takes time to generate on the test set with a beam size of $5$.}
	\label{tab:efficiency}
\end{table}

\subsection{Efficiency Comparison}
In addition to accurately reproducing results, we have optimized TextBox 2.0 for computational efficiency. We streamline the training process and support efficient decoding strategies. To compare the efficiency, we choose the well-known PLM libraries Fairseq\footnote{We utilize the code from Fairseq 0.12.2.} and Hugging Face\footnote{We utilize the code from Transformers 4.20.1.}, and then test the time consumption under identical settings described in Section~\ref{sec:res}.

From the results in Table~\ref{tab:efficiency}, we can see that our TextBox 2.0 is more efficient than Fairseq and Hugging Face. During training, TextBox 2.0 simplifies the training process and reduces the time spent on non-essential functions such as trainer management and loss tracking. In the generation process, our library is significantly faster than the other two libraries due to the incorporation of efficient decoding strategies introduced in Section~\ref{sec-training}.

\subsection{Visualization Analysis} \label{sec:visual}
Besides reproducing a model, it is also important to compare existing methods, analyze the generated texts, and explore directions for improvement. Our library sets a specific leaderboard for each dataset, including basic metric results, author repositories, and generated texts. Figure~\ref{fig:analysis} (a) showcases the leaderboard for the CNN/DailyMail dataset.

Users can also utilize TextBox 2.0 to conduct visualization analysis for specified models. For example, our library can automatically plot the boxplot of the ROUGE-L score for different input lengths and the $n$-gram overlap of target and generated texts with the input document. From the results in Figure~\ref{fig:analysis} (b), we can find that T5 excels at short document summarization while BART excels at long document summarization.
It is useful to analyze and improve the deficiencies of text generation models or obtain better performance by combining their results.
As another example, Figure~\ref{fig:analysis} (c) illustrates that BART and T5 have a significantly higher $n$-gram overlap ratio than golden sentences, indicating that they tend to ``copy'' the input document rather than ``summarize'' it. From such analysis results, users can apply the methods proposed by~\citet{goyal-etal-2022-training} to alleviate it.